\documentclass{article}
\usepackage{spconf,amsmath,graphicx}
\usepackage{bm}
\usepackage{tablefootnote}
\usepackage{pdfpages}
\usepackage[numbers, comma, sort&compress]{natbib}

\DeclareMathAlphabet\mathbfcal{OMS}{cmsy}{b}{n}
\usepackage{amssymb}  

\DeclareMathAlphabet      {\mathbfit}{OML}{cmm}{b}{it}

\title{Some New Layer Architectures for Graph CNN}
\name{Shrey Gadiya{*}\thanks{{*} Joint co-first authors.}, \qquad Deepak Anand{*}, \qquad Amit Sethi\thanks{This work was supported by Nvidia GPU grant.}}
\address{Department of Electrical Engineering,\\
    Indian Institute of Technology Bombay,\\
    Mumbai 400076.}
\begin{document}
%
\maketitle
\begin{abstract}
While convolutional neural networks (CNNs) have recently made great strides in supervised classification of data structured on a grid (e.g. images composed of pixel grids), in several interesting datasets, the relations between features can be better represented as a general graph instead of a regular grid. Although recent algorithms that adapt CNNs to graphs have shown promising results, they mostly neglect learning explicit operations for edge features while focusing on vertex features alone. We propose new formulations for convolutional, pooling, and fully connected layers for neural networks that make more comprehensive use of the information available in multi-dimensional graphs. Using these layers led to an improvement in classification accuracy over the state-of-the-art methods on benchmark graph datasets.
\end{abstract}
\section{Introduction}
\label{sec:intro}
In many applications, we encounter data whose features are related to each other as a general graph as opposed to a regular grid. For example, one can represent molecular structures as a graph, with vertices representing atoms and edge features characterizing the bonds between individual atoms. Inspired by the success of convolutional neural networks (CNNs) on classifying grid structured data such as images and speech, graph CNN architectures for classifying graphs have recently been proposed ~\cite{DBLP:journals/corr/DefferrardBV16, 2012arXiv1206.6483K, DBLP:journals/corr/SuchSDPZMCP17,DBLP:journals/corr/BrunaZSL13,DBLP:journals/corr/abs-1711-08920,DBLP:journals/corr/KipfW16,pmlr-v48-niepert16,DBLP:journals/corr/PerozziAS14}. 

Most prior works on graph CNNs have concentrated on filtering vertex attributes or labels, while edge attributes or labels have received little attention. For instance, a method that we shall call robust spatial filtering (RSF) proposed filtering and pooling operations that explicitly modify vertex features~\cite{DBLP:journals/corr/SuchSDPZMCP17}. However, the layer architectures used by RSF have several restrictions such as a complete dependence of edge feature transformations on vertex features in the convolutional and fully connected layers. Especially for datasets with rich edge features, such restrictions can lead to lower classification accuracy.

Our work significantly extends RSF in three ways. Firstly, we propose a novel convolution operation with explicit learnable parameters for not only vertex features but also edge features. The proposed operation also does away with the restriction that the resultant adjacency matrix has to be symmetric. Secondly, we propose modifications to the higher pooling layers that also relax the need for the resultant adjacency matrix to be symmetric without significantly increasing the number of parameters. Lastly, we also propose a method to use edge features in the fully connected layers of a neural network. Each of the proposed layer architectures independently led to improvements in graph classification accuracy on benchmark datasets.

\section{Background and Related Work}
\label{sec:related}

We start with describing a general graph structure that has both vertex and edge attributes (or features). Formally, a multi-attributed graph with $N$ vertices can be represented as $\bm{G}=(\bm{V},\mathbfcal{A})$, where $\bm{V}\in \mathbb{R}^{N\times F}$ is the vertex feature matrix in which each of the $N$ vertices has $F$ features (or attributes). Here, $\mathbfcal{A}\in\mathbb{R}^{N\times N \times L}$ is the adjacency (or edge feature) tensor in which each of the $N^2$ edges is associated with a feature or attribute vector of size $L$. Normal definitions hold such as $a_{ijl}=0$ implies the absence of an edge in the $l^{th}$ slice between $i^{th}$ and $j^{th}$ vertices.

Deep learning approaches for graph-structured data can be divided into two categories -- spectral approaches that use the graph Fourier transform, and spatial approaches that aggregate features from a spatial neighborhood. 

\subsection{Spectral approaches}
The first set of approaches are based on spectral graph theory. Consider a graph $\bm{G}$ with a single adjacency matrix $\bm{A}_1\in \mathbb{R}^{N\times N}$ and a graph signal $\bm{x} \in \mathbb{R}^{N\times 1}$. Here, $\bm{x}$ can be considered as a vertex matrix with $F=1$. Let $\bm{D}$ be its degree matrix such that:
\begin{align}
    \bm{D}_{ii}=\sum_{j=1}^{N}\bm{A}_{1}(i,j)
\end{align}
Then the normalized graph Laplacian is defined as:
\begin{gather}
    \bm{L} =\bm{I} - \bm{D}^{-\frac{1}{2}}\bm{A}_1\bm{D}^{-\frac{1}{2}}
\end{gather}
The eigenbasis matrix $\bm{U}$ of $\bm{L}$ is then used for graph Fourier transform (GFT). Note that if $\bm{A}_1$ is symmetric (i.e., the data is modeled as undirected graphs) then $\bm{U}$ is orthonormal.

Spectral approaches involve learning a frequency domain filter $\bm{h}$ for transforming the graph signal $\bm{x}$ as follows: 
\begin{align}
    \bm{x}*\bm{h} = \bm{U}^T\left(\left(\bm{Ux}\right)\odot\bm{h}\right)
\end{align}
where $*$ represents filtering and $\odot$ represents element-wise multiplication.

A major disadvantage of the spectral approaches is that the GFT basis depends on the graph structure, in particular, the adjacency matrix. Hence these approaches work only on homogeneous graphs where each graph can differ only in the vertex features while the adjacency matrix should remain constant. Moreover, these approaches often require the adjacency matrix to be symmetric, which holds only for undirected graphs. Another disadvantage of these methods is the absence of localized filtering or feature aggregation. Although recent works in this direction solve some of these problems using localized spectral filtering and polynomials of graph Laplacian~\cite{DBLP:journals/corr/BrunaZSL13}, \cite{DBLP:journals/corr/DefferrardBV16}, \cite{DBLP:journals/corr/KipfW16}. 

\subsection{Spatial approaches}
Spatial approaches do not rely on spectral graph theory to perform filtering. Instead, different architectures define different ways to learn to combine the information from neighboring vertices in the form of local filters. For example, PATCHYSAN transforms the graph data into a grid structure using graph normalization schemes to sort the vertices~\cite{pmlr-v48-niepert16}. This grid structure can be then passed to a traditional CNN. DeepWalk uses random walks to perform localized learning over the neighborhood~\cite{DBLP:journals/corr/PerozziAS14}. SplineCNN learns a spline function using the B-spline basis to sample convolutional weights based on the relative pseudo-coordinates of the neighboring vertices~\cite{DBLP:journals/corr/abs-1711-08920}. 

Our work is based on RSF, which is a spatial domain architecture that proposed a graph filter for the convolution layer and a pooling operation known as graph embed pooling (GEP) to downsize a variable-sized input graph to a fixed-sized output graph~ \cite{DBLP:journals/corr/SuchSDPZMCP17}. Among the other methods mentioned above, this is the only method that can learn general graph transformations across its architecture due to its elegant design for filtering graphs. Most likely, due to its emphasis on learning to compute graphs from input graphs in each layer, it also gives higher classification accuracy on benchmark datasets.

The graph filter (convolutional layer) proposed in RSF filters a vertex matrix $\bm{V}_{in}\in\mathbb{R}^{N\times F}$ using an adjacency tensor $\mathbfcal{A}\in\mathbb{R}^{N\times N\times L}$ to compute an output vertex matrix $\bm{V}_{conv}\in\mathbb{R}^{N\times 1}$ as follow:
\begin{gather}
    \bm{V}_{conv} = \sum_{f=1}^F \bm{H}^{(f)}\bm{V}_{in}^{(f)}+b \\
    \bm{H}^{(f)} = h_0^{(f)}\bm{I} + \sum_{l=1}^L \bm{A}_l h_l^{(f)}
\end{gather}
where $\bm{H}\in\mathbb{R}^{N\times N\times F}$ and $h_i^{(f)}$'s are learnable parameters. In general, one uses multiple filters to obtain $F'$ output features. That is, $\bm{V}_{conv}\in\mathbb{R}^{N\times F'}$ is computed using tensor $\bm{H}\in\mathbb{R}^{N\times N\times F'\times F}$ as follows:
\begin{align}
    \bm{V}_{conv}^{(f')} = \sum_{f=1}^F \bm{H}^{(f,f')}\bm{V}_{in}^{(f)}+b^{(f')}  \label{V_conv}
\end{align}

The graph embed pooling (GEP) proposed in RSF uses an embedding matrix $\bm{V}_{emb}\in\mathbb{R}^{N\times N'}$ to obtain a graph of fixed $N'$ number of vertices from a graph of any size with $F$ vertex features~\cite{DBLP:journals/corr/SuchSDPZMCP17} as follows:
\begin{equation}
    \bm{V}_{emb}^{(f')} = \text{softmax}(\bm{V}_{conv}^{(f')}) \label{eq:Vemb}
\end{equation}
The outputs $\bm{V}_{out}$ and $\bm{A}_{out}$ are given by:
\begin{gather}
    \bm{V}_{out}=\bm{V}_{emb}^{T}\bm{V}_{in}\label{poolV}\\
    \bm{A}_{out}=\bm{V}_{emb}^{T}\bm{A}_{in}\bm{V}_{emb} \label{PoolA}
\end{gather}

There are three main limitations of the RSF architecture~\cite{DBLP:journals/corr/SuchSDPZMCP17}. Firstly, there is no explicit convolutional filtering for the edge features. There is only implicit communication between the edge and the vertex features in each convolutional layer as per equation~\ref{PoolA}. The edge features may also contain substantial information independent of the vertex features and there may be a benefit of filtering them separately. Examples of such datasets include the BZR\_MD, COX2\_MD, ER\_MD, and DHFR\_MD from~\cite{KKMMN2016}. In these datasets, the vertex labels represent the atom type and the edge features encode various distances between atoms defined in section 3.2 of ~\cite{2012arXiv1206.6483K}. Secondly, RSF uses only flattened vertex features before the final layer for graph classification. Edge features are, once again, ignored after the last pooling/convolution layer and do not explicitly contribute to the final classification. Lastly, as per (\ref{poolV}) and (\ref{PoolA}), if the input graph adjacency matrix is symmetric, the symmetricity is further propagated throughout by the pooling layers, which seems like an unnecessary restriction.

\section{Proposed Architecture}

Based on an analysis of the limitations of RSF's layer architectures presented in the previous section, we propose modifications to all the three types of layers -- convolutional, pooling, and fully connected.

\subsection{Edge features convolutional layers}
\label{sec:convProposed}
Assuming a general graph structure that has edge features, we propose a new formulation for the convolutional layer that outputs an adjacency tensor $\mathbfcal{A}_{out} \in \mathbb{R}^{N\times N\times L'}$, where $L'$ is the number of features per edge output by this layer that uses vertex features $\bm{V}_{in}$ and edge features $\mathbfcal{A}_{in}$ as inputs. The $k^{th}$ output channel $\mathbfcal{A}_{out}(:,:,k)$ is an adjacency matrix just like the $l^{th}$ input channel $\mathbfcal{A}_{in}(:,:,l)$ is. We wanted the edge convolutional function to have the following properties. Firstly, $\mathbfcal{A}_{out}(i,j,k)$ should depend on $\mathbfcal{A}_{in}(i,j,:)$, i.e., on all the channels of the input edge. Secondly,  $\mathbfcal{A}_{out}(i,j,k)$ should depend on the vertex features $\quad \bm{V}(i,:)$ and $\bm{V}(j,:)$.

\begin{figure}
    \centering
   \includegraphics[width=1\linewidth]{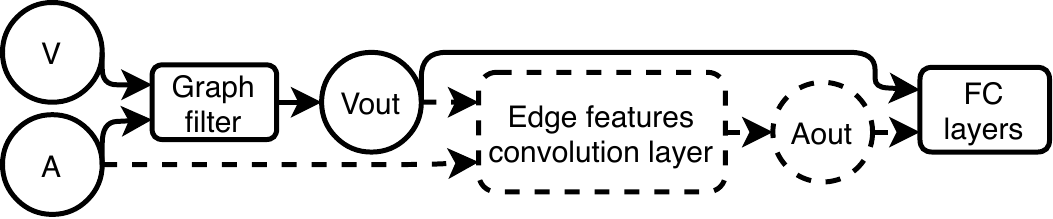}
    \caption{RSF convolutional architecture~\cite{DBLP:journals/corr/SuchSDPZMCP17} (solid) with the proposed modification for edge features shown (dotted).}
    \label{fig:flow}
\end{figure}

 To define an edge convolution (as shown using the dotted structure in figure~\ref{fig:flow}, for an edge between vertices $i$ and $j$ we stack the input edge feature $\mathbfcal{A}_{in}(i,j,:)$ and input vertex features $\bm{V}_{in}(i,:)$ and $\bm{V}_{in}(j,:)$ to form a $L_{in} + 2F$ dimensional vector $\bm{X}_{ij}$. We then learn a filter weight matrix $\bm{W}\in\mathbb{R}^{L_{out}\times (2F+L_{in})}$ that performs the following operation:
\begin{align}
    \mathbfcal{A}_{out}(i,j,:) =  \phi(\bm{WX}_{ij}) 
\end{align}
where $\phi$ is a monotonic nonlinear activation function. 

We encountered spikes in the loss function while training when sigmoid and ReLU activation functions were used. To remedy this, we experimented with different activations functions, and found $\tanh{(\mathrm{ReLU}(.))}$ to give a well-behaved decay in loss. The added advantage of this activation is that its output is in the range $[0,1)$, where $0$ represents the absence of an edge. A sharp change in the gradient at $0$ input ensures that several edges are eliminated (sparseness), which a sigmoid function cannot ensure with its range being $(0,1)$. Moreover, this activation function maps large positive inputs smoothly and asymptotically to $1$, while a non-zero gradient for a high input ensures learnability unlike truncated ReLU. 

\subsection{Pooling layers}
\label{cite:poolProposed}
Pooling operations transform an input graph of size $N$ to an output graph of a smaller size $N’$ with adjacency tensor $\mathbfcal{A}_{out}\in \mathbb{R}^{N'\times N'\times L}$ and vertex matrix $\bm{V}_{out}\in \mathbb{R}^{N'\times F}$. We propose variants for two different types of pooling layers.

\subsubsection{Global level pooling}
\label{GEPAsym}
Pooling is a way to derive salient vertices (or super-vertices) by taking a weighted combination of the underlying vertices and their edges. Consider the following pooling layer defined in ~\cite{DBLP:journals/corr/abs-1803-10071}:
\begin{gather}
    \mathbfcal{A}_{out}(:,:,i) = \bm{K}^T\mathbfcal{A}_{in}(:,:,i)\bm{K}\\
    \bm{V}_{out} = \bm{K}^T\bm{V}_{in}
\end{gather}
where $\bm{K}\in\mathbb{R}^{N\times N'}$ is a learnable weight matrix. We refer to this as global level pooling (GLP). 

First, we propose to generalize GLP as follows:
\begin{gather}
    \mathbfcal{A}_{out}(:,:,i) = \bm{K}_1^T\mathbfcal{A}_{in}(:,:,i)\bm{K}_2\\
    \bm{V}_{out} = \bm{K}_3^T\bm{V}_{in}
\end{gather}
Then, we propose two variants of this generalization:
\begin{itemize}
    \item \textbf{GLPAsym:} To allow the resultant adjacency matrix to be asymmetrical (transform both directed and undirected graph to directed graph), we allow $\bm{K}_1\neq \bm{K}_2\neq \bm{K}_3\neq \bm{K}_1$. Experimentally, better results are obtained if weights in $\bm{K}_1$ and $\bm{K}_2$ are initialized with the same values, although they may evolve during training to have different values. 
    \item \textbf{GLPSym:} To obtain a symmetrical resultant adjacency matrix, we constrain $\bm{K}_1= \bm{K}_2$ and allow $\bm{K}_1\neq \bm{K}_3$.  
\end{itemize}

One remaining limitation of GLP is that the input graph is required to be of a fixed size $N$. Therefore, RSF uses graph embed pooling (GEP). We use GEP as the first pooling layer in our experiments to obtain a graph with a fixed number of vertices before experimenting with GLP.

\subsubsection{Graph embed pooling}
In graph embed pooling (GEP), the generic matrices $\bm{K}_1,\bm{K}_2$, $\bm{K}_3$ are replaced with $\bm{V}_{emb1},\bm{V}_{emb2},\bm{V}_{emb3}$ respectively, where the latter three are defined as per equation~\ref{eq:Vemb}. Hence, similar to the above variants for GLP, we propose the use of GEPAsym and GEPSym. Our experiments show that these variants perform better than the original GEP proposed in~\cite{DBLP:journals/corr/SuchSDPZMCP17}.

\subsection{Fully connected layers for edge features}
\label{afc}
To get the full benefit of the edge features, we propose a new formulation for the fully connected (FC) layers also. Specifically, to use the adjacency tensor $\mathbfcal{A}_{in}$ obtained by the last convolutional or pooling layer, we reshape it to an $N\times NL$ matrix and concatenate it to the vertex matrix $\bm{V}_{in}$ to obtain a $N\times (F+NL)$ matrix $\bm{Y_{in}}$. In other words, unlike in RSF where only the vectorized vertex matrix was fed to the FC layers, we additionally pass the vectorized adjacency tensor.

\section{Experiments and Results}
We performed experiments on several benchmark datasets to test the significance of each of the three proposed layer architectures. Notations for the architectures are given in table~\ref{tab:arch}. 

\begin{table}
  \centering
    \caption{Notation for architectural description}
    \label{tab:arch}
    \begin{tabular}{|c|l|} 
      \hline
      \textbf{Symbol} &   \textbf{Interpretation} \\
      \hline
      $n$F   & Graph filter with $n$ output vertex features \\ \hline
      $n$EF  & Edge features convolutional layer with $n$ \\
      & output edge features\\ \hline
      P$n$  & Graph pooling layer with output graph size $n$\\ \hline
      EFC$n$& Fully connected layer with input as defined in\\
      & section \ref{afc} and having $n$ output features\\ \hline
      FC$n$  & Fully connected layer with $n$ output features\\ \hline
    \end{tabular}
\end{table}

\subsection{Adding convolution layer for edge features}

Table \ref{tab:edge_conv} establishes the usefulness of the proposed convolutional layers that explicitly learn weights to transform edge features over the previous state of the art~\cite{2012arXiv1206.6483K} on datasets with rich edge features -- BZR\_MD, COX2\_MD, ER\_MD and DHFR\_MD~\cite{KKMMN2016}. Due to the small sizes of the datasets, our results are based on five-fold cross-validation.

The use of the proposed edge features convolutional layers do not always guarantee an improvement in graph classification accuracy, however, especially in cases when the edge features do not have significant information. For example, we saw a small drop in classification accuracy over the NCI1 dataset, which has three binary adjacency matrices, when we used the proposed edge convolutional layers.

\begin{table}
  \centering
    \caption{Comparison of graph classification accuracy (mean$\pm$std. dev. in \%) using different architectures (where, ours used the proposed GEPAsym pooling layer in a 2F-7EF-4F-6EF-P32-3F-4EF-P8-EFC280 architecture).}
    \label{tab:edge_conv}
    \begin{footnotesize}
    \begin{tabular}{|c|r|r|r|r|} 
      \hline
      \textbf{Network} & \textbf{DHFR\_MD} & \textbf{ER\_MD} & \textbf{BZR\_MD} & \textbf{COX2\_MD} \\
      \hline
      Ours  & \textbf{83.1 $\pm$ 2.2} &\textbf{83.8 $\pm$ 1.7 }&\textbf{80.0 $\pm$ 2.8} & 74.3 $\pm$ 3.9\\ \hline
      (C)SM \cite{2012arXiv1206.6483K}  &79.9 $\pm$ 1.1 &82.0 $\pm$ 0.8 &79.4 $\pm$ 1.2 &74.4 $\pm$ 1.7\\ \hline
      PH \cite{2012arXiv1206.6483K}   &80.8 $\pm$ 1.2 &81.4 $\pm$ 0.6&77.9 $\pm$ 1.6 & \textbf{74.6 $\pm$ 1.5}\\ \hline
      FLRW \cite{2012arXiv1206.6483K}   &79.1 $\pm$ 1.1 &81.6 $\pm$ 1.1&77.9 $\pm$ 1.1 &74.4 $\pm$ 1.5\\ \hline
      SP \cite{2012arXiv1206.6483K}   &77.6 $\pm$ 1.5 &79.9 $\pm$ 1.3&78.2 $\pm$ 1.2 &74.5 $\pm$ 1.3\\
      \hline
    \end{tabular}
    \end{footnotesize}
\end{table}


\subsection{Changing the second pooling layer}

To test the efficacy of the proposed pooling layer, we reproduced the architectures mentioned in \cite{DBLP:journals/corr/SuchSDPZMCP17} for D\&D and NCI1 datasets and replaced the second pooling layer with different configurations of the GLP and GEP layer proposed in Section~\ref{cite:poolProposed}. We did not change the first pooling layer since GLP requires the graph to have a fixed number of vertices. Additionally, changing the first pooling layer from the original GEP~\cite{DBLP:journals/corr/SuchSDPZMCP17} to the ones that we proposed did not improve the accuracy much. The architecture for NCI1 was 2$\times$64F-P32-32F-P8-FC256 and for D\&D was 2$\times$64F-P32-64F-P8-FC256, where we changed the P8 layer (the P32 layer being fixed to GEP). The results are shown in table~\ref{tab:pool2}.
 
\begin{table}
  \begin{center}
    \caption{Comparison of graph classification accuracy (mean$\pm$std. dev. in \%) using different pooling layer architectures by keeping the number of total parameters in a tight range (118,452 to 121,332 for NCI1, and 165,596 to 166,476 for D\&D).}
    \label{tab:pool2}
    \begin{tabular}{|c|r|r|} 
      \hline
      \textbf{Second pooling layer} & \textbf{NCI1} & \textbf{D\&D} \\
      \hline
      GEP (RSF)  & 84.6 $\pm$ 2.2 & 81.9 $\pm$ 3.4\\ \hline
      GEPAsym & 85.0 $\pm$ 1.5 & 81.8 $\pm$ 2.4 \\ \hline
      GEPAsym\tablefootnote{Symmetric weight initialization was used as mentioned in section \ref{GEPAsym}.} & 84.8 $\pm$ 1.1 & 81.9 $\pm$ 2.7\\ \hline
      GEPSym & 84.4 $\pm$ 0.8 & 81.8 $\pm$ 2.4 \\ \hline
      GLP  & 85.1 $\pm$ 1.6 & 81.9 $\pm$ 3.1 \\ \hline
      GLPAsym & 85.0 $\pm$ 1.9 & \textbf{82.6 $\pm$ 2.7} \\ \hline
      GLPAsym\footnote[1] & \bf{85.2 $\pm$ 1.0} & 82.4 $\pm$ 2.4 \\ \hline
      GLPSym & 84.8 $\pm$ 1.5 & 82.3 $\pm$ 3.2 \\ \hline
    \end{tabular}
  \end{center}
\end{table}


\subsection{Adding fully connected layer for edge features}

Results from using the two variants of a fully connected layer that incorporates edge features as proposed in section~\ref{afc} are shown in table~\ref{tab:afc}, where ten-fold cross validation was used. Compared to the previous architectures, our last FC layers have more input features.

\begin{table}
  \begin{center}
    \caption{Results for NCI1 and D\&D datasets: using flattened adjacency tensor in the fully connected layers. Accuracies are in \% with mean and standard deviation}
    \label{tab:afc}
    \begin{footnotesize}
    \begin{tabular}{|l|c|r|} 
      \hline
      \textbf{Dataset} &\textbf{Architecture} & \textbf{Accuracy} \\
      \hline
      NCI1 & 2$\times$64F-P32-32F-P8-32F-EFC448  & 85.2 $\pm$ 1.3 \\ \cline{2-3}
       &2$\times$64F-Pool32-64F-P8-32F-EFC448-FC32  & \textbf{85.4 $\pm$ 1.4}   \\
      \hline
      D\&D & 2$\times$64F-P32-32F-P8-64F-EFC576  & 79.7 $\pm$ 3.3 \\ \cline{2-3}
       &2$\times$64F-Pool32-64F-P8-64F-EFC576-FC32  & 81.7 $\pm$ 2.9\\
       \hline
    \end{tabular}
    \end{footnotesize}
  \end{center}
\end{table}

\section{Conclusion and discussion}
In this paper we proposed new formulations for graph convolutional, pooling and fully connected layers that explicitly incorporate learned transformations of edge features. We explored how one can have more generality in a graph CNN to improve performance in different cases. Our experiments shows that using these layers lead to the performance gains in datasets with rich and multi-dimensional edge features as is common in nature and complex systems. Extensions of our work can lead to a better understanding of such complex interactions. 

\bibliographystyle{IEEEbib}

\bibliography{refs}

\end{document}